%% file: main.tex
\documentclass[letterpaper]{article}
\usepackage[preprint]{aaai2027}
\usepackage[hyphens]{url}
\usepackage{graphicx}
\usepackage{natbib}
\usepackage{caption}
\usepackage{amsmath,amssymb}
\usepackage{algorithm}
\usepackage{algorithmic}
\usepackage{booktabs}

\newcommand{\ee}[1]{\ensuremath{\times 10^{#1}}}

\title{Perturbative-NeuSA: A Structured Spectral Framework\\
for Time-Dependent PDEs}
\author{Xianli Zhu, Jia Yin\corresponding}
\affiliations{
School of Mathematical Sciences, Fudan University\\
Shanghai, China\\
\texttt{25210180115@m.fudan.edu.cn}, \texttt{jiayin@fudan.edu.cn}
}

\begin{document}
\maketitle

\input{src/abstract}
\input{src/introduction}
\input{src/related_work}
\input{src/method}
\input{src/experiments}
\input{src/discussion}
\input{src/conclusion}

\bibliography{references}

\clearpage
\raggedbottom
\setcounter{table}{0}
\setcounter{figure}{0}
\setcounter{equation}{0}
\begin{center}
{\LARGE\bf Supplementary Material}
\end{center}

\input{src/supp_derivations}
\input{src/supp_protocol}
\input{src/supp_results}
\input{src/supp_reproducibility}
\input{src/supp_visualizations}

\end{document}

%% file: src/abstract.tex
\begin{abstract}
Neural spectral PDE solvers often learn an entire unresolved vector field even when an inexpensive approximate model can already capture most of the trajectory. Here we introduce Perturbative-NeuSA, a residual formulation that decomposes the target solution into a low-fidelity background and a high-resolution perturbation, so that only the unresolved dynamics is learned. Starting from the exact perturbation equation, the method combines a fixed spectral operator, a background-dependent correction, the background defect in the target PDE, and an optional neural closure. This construction makes the roles of physical structure and neural closure separately measurable. Across 2D Burgers, Klein--Gordon, and heterogeneous 2D wave equations, the deterministic structured solver outperforms the trained NeuSA baseline while requiring no neural-network training. The largest gains occur on Burgers, where the deterministic correction reduces training and extrapolation errors by factors of 24 and 44, respectively.
In addition, a Klein--Gordon sweep over seven background resolutions shows that the effect of the closure is conditional: it improves a poor background by 3.6 times, becomes neutral at intermediate resolutions, and degrades a well-resolved background. For the wave equation, however, the closure provides an additional 18\% reduction when the remaining residual is interface-localized. Multi-initial-condition diagnostics further show that the useful closure regime depends on the initial-condition spectrum and can disappear in extrapolation when structured correction already captures the dominant Burgers dynamics. Perturbative-NeuSA therefore reframes neural closure as a conditional, diagnosable correction governed by background fidelity, residual organization, and compatibility with the closure model.
\end{abstract}

%% file: src/introduction.tex
\section{Introduction}

Physics-informed neural networks (PINNs) enforce partial differential equations through a training objective rather than a conventional time-stepping scheme~\cite{raissi2019physics}. Their mesh-free representation is flexible, but in time-dependent problems these methods remain vulnerable to low-frequency bias~\cite{xu2019frequency,wang2022when}, violations of temporal causality~\cite{wang2024respecting}, and poor extrapolation beyond the training interval~\cite{kim2021dpm}. Related neural-operator models also face out-of-distribution extrapolation challenges~\cite{zhu2023reliable}. Neuro-Spectral Architectures (NeuSA) address part of this difficulty by projecting the target PDE onto a spectral basis and evolving the resulting coefficients with a Neural ODE~\cite{chen2018neural,bizzi2025neusa}. A linearized spectral operator initializes the coefficient vector field, while a multilayer perceptron learns the nonlinear, heterogeneous, or otherwise unresolved dynamics beyond that prior.

However, for nonlinear equations, this residual can still contain terms that
are already identifiable from the PDE and a coarse trajectory. NeuSA's fixed
prior only contains a state-independent linear operator, leaving the network
to reconstruct the full state-dependent residual. Classical defect-correction
methods take a different route: they compute an inexpensive approximate trajectory, evaluate its defect in the target equation, and solve for a correction~\cite{bohmer1984defect}.
This motivates a narrower role for the neural closure. The analytically
identifiable residual should be exposed first, while a neural closure should
model only the terms that remain unresolved.

\paragraph{Motivation.}
If an inexpensive background already tracks most of the trajectory, training a
neural closure on the full vector field unnecessarily spends model capacity on terms that
can be computed explicitly. To address this, we decompose
$h=h_{\mathrm{bg}}+\widetilde h$, represent the background defect explicitly,
and restrict the learned target to the higher-order and otherwise unresolved
perturbation terms. We instantiate this idea on three PDEs:
\begin{align}
\text{2D Burgers:}\quad
&\partial_t\boldsymbol q=\nu\Delta\boldsymbol q
-(\boldsymbol q\cdot\nabla)\boldsymbol q,\nonumber\\
\text{Klein--Gordon:}\quad
&u_t=v,\qquad v_t=u_{xx}-m\sin u,\nonumber\\
\text{Heterogeneous wave:}\quad
&u_t=v,\qquad v_t=c^2(y)\Delta u. \nonumber
\end{align}
For the Klein--Gordon equation, for example,
expanding $-m\sin u$ around the background gives the first-order term
$-m\cos(u_{\mathrm{bg}})\widetilde u$. After separating the fixed linear term
$-m\widetilde u$, the background-dependent correction becomes
$-m[\cos(u_{\mathrm{bg}})-1]\widetilde u$.

\begin{figure*}[t]
    \centering
    \includegraphics[width=\textwidth]{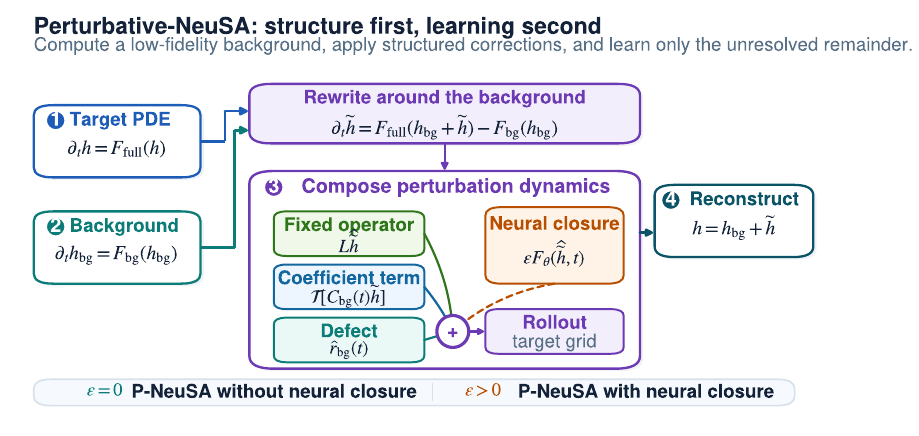}
    \caption{Perturbative-NeuSA separates a low-fidelity background solve from a high-resolution perturbation solve. Setting $\varepsilon=0$ gives a deterministic structured solver. Setting $\varepsilon>0$ adds a learned residual correction.}
    \label{fig:overview}
\end{figure*}

We implement this principle by decomposing the full state as
\begin{equation}
    h(t)=h_{\mathrm{bg}}(t)+\widetilde h(t),
    \label{eq:decomposition}
\end{equation}
where $h_{\mathrm{bg}}$ is generated by a low-fidelity background operator and $\widetilde h$ is a perturbation represented at the target spectral resolution. The resulting perturbation equation separates three structured contributions: a fixed linear operator, a background-dependent correction, and the defect of the background in the target PDE. An optional neural term with weight $\varepsilon$ models higher-order or unresolved effects. We call this structured perturbative framework Perturbative-NeuSA (P-NeuSA). Figure~\ref{fig:overview} summarizes its construction.

This hierarchy makes the ablation directly interpretable. The gap between the background
and P-NeuSA with $\varepsilon=0$ measures the contribution of the analytical correction, whereas
the gap between P-NeuSA with $\varepsilon=0$ and P-NeuSA with $\varepsilon>0$ measures the
marginal value of the neural closure.

This work makes three contributions:
\begin{itemize}
    \item First, we propose P-NeuSA to reduce the burden of learning the full state-dependent PDE residual by separating a low-fidelity background, a structured high-resolution correction, and an optional learned closure. The formulation covers nonlinear truncated backgrounds for Burgers and Klein--Gordon and a smoothed-coefficient background for heterogeneous wave propagation.
    \item Second, we show that P-NeuSA can outperform NeuSA even in its deterministic form, before adding any neural closure ($\varepsilon=0$). Across three PDEs, the largest improvements occur for the Burgers equation, where P-NeuSA reduces training and extrapolation errors by factors of 24 and 44, respectively.
    \item Third, rather than assuming that neural closure is uniformly beneficial, we identify when it is beneficial through resolution, multi-initial-condition, and residual-localization studies, showing that closure gains depend on the spectrum resolved by the background and the organization of the remaining residual, while the Burgers ablation reveals structure-dominated regimes in which the tested closure is not consistently beneficial.
\end{itemize}

%% file: src/related_work.tex
\section{Related Work}

\paragraph{Physics-informed neural solvers.}
PINNs represent a solution with a coordinate network and penalize violations of
the governing equation~\cite{raissi2019physics}. Gradient-enhanced objectives,
quadratic residual networks, sinusoidal features, transformers, and causal
training improve particular failure modes~\cite{bu2022quadratic,yu2022gradient,
zhao2023pinnsformer,wang2024respecting,wong2024learning}. Nevertheless, these approaches still optimize a global function approximator for the solution or its residual. In contrast, our method preserves an explicit time integrator and measures the
marginal value of a learned vector-field correction after a low-fidelity
trajectory and its analytical correction are provided.

\paragraph{Spectral and Neural ODE formulations.}
Neural spectral methods shift the learning problem to basis coefficients, where
spatial derivatives become structured algebraic operators~\cite{du2024neural}.
NeuSA combines this representation with Neural ODE integration and a fixed
linear prior~\cite{chen2018neural,bizzi2025neusa}. Our method retains
NeuSA's causal coefficient dynamics and fixed linear prior, and augments them
with a task-specific background trajectory, its defect, and a
background-dependent correction. The Jacobian along this trajectory is
evaluated pseudo-spectrally, shifting the learned target from the full
state-dependent residual to the unresolved higher-order or coefficient-mismatch
terms left after the structured perturbation correction.

\paragraph{Classical correction and reduced models.}
Perturbation expansions, multigrid, preconditioning, and projection-based
model reduction share the broader strategy of replacing a difficult problem
with a structured approximation or an easier auxiliary problem~\cite{bender1999advanced,briggs2000multigrid,saad2003iterative,
benner2015survey}. Our formulation follows the same structural principle, but
targets a time-dependent spectral Neural ODE. Empirically, we introduce a four-level comparison among the background solver, structured perturbative correction, learned closure, and NeuSA, together with resolution and multi-initial-condition studies that identify when learning the remaining correction is beneficial.

%% file: src/method.tex
\section{Perturbative Spectral Formulation}
\label{sec:method}

\paragraph{Spectral state.}
We represent each PDE by a first-order state $h$. For second-order equations, $h=(u,v)$ with $v=\partial_t u$. Let $\mathcal{T}$ map a physical-space field to its retained spectral coefficients, $\widehat h=\mathcal{T}h$, and let $\mathcal{T}^{-1}$ denote the inverse transform~\cite{boyd2001chebyshev}. For basis functions $\{b_k\}_{k=1}^{M}$,
$h(t,x)\approx\sum_{k=1}^{M}\widehat h_k(t)b_k(x)$.
Here $x$ is the spatial coordinate, $k$ indexes retained modes, and $\{\widehat{h}_k\}_{k=1}^{M}$ denote
spectral coefficients.
NeuSA integrates
\begin{equation}
    \frac{d\widehat h}{dt}=\widehat F_0(\widehat h)
    +\varepsilon F_\theta(\widehat h,t),
    \label{eq:neusa}
\end{equation}
where $\widehat F_0$ is a fixed linear prior, $F_\theta$ is a neural vector field with parameters $\theta$, and $\varepsilon\ge0$ is the neural-closure weight.

\paragraph{Low-fidelity background operator.}
Let $F_{\mathrm{full}}$ be the target first-order PDE operator. We first compute a background trajectory
\begin{equation}
    \partial_t h_{\mathrm{bg}}=F_{\mathrm{bg}}(h_{\mathrm{bg}}),
    \label{eq:background}
\end{equation}
where $F_{\mathrm{bg}}$ is chosen to be cheaper, smoother, or lower-resolution than $F_{\mathrm{full}}$.

\paragraph{Exact perturbation equation.}
Define the perturbation by $\widetilde h=h-h_{\mathrm{bg}}$. Differentiating this identity and substituting $\partial_t h=F_{\mathrm{full}}(h)$ and $\partial_t h_{\mathrm{bg}}=F_{\mathrm{bg}}(h_{\mathrm{bg}})$ into Eq.~\eqref{eq:decomposition} gives the exact perturbation equation
\begin{equation}
    \partial_t\widetilde h
    =F_{\mathrm{full}}(h_{\mathrm{bg}}+\widetilde h)
    -F_{\mathrm{bg}}(h_{\mathrm{bg}}).
    \label{eq:exact-perturbation}
\end{equation}
Thus the full PDE is rewritten in terms of the perturbation \(\widetilde h\), where the background supplies the reference trajectory, and the perturbation equation corrects both the background defect and the unresolved high-resolution dynamics.
Expanding $F_{\mathrm{full}}(h_{\mathrm{bg}}+\widetilde h)$ around $h_{\mathrm{bg}}$ yields
\begin{equation}
    \partial_t\widetilde h
    =L\widetilde h+C_{\mathrm{ex}}(t)\widetilde h
    +r_{\mathrm{bg}}(t)+R_2(t,\widetilde h),
    \label{eq:perturbation}
\end{equation}
where $L$ denotes the state-independent linear operator of the target PDE, which NeuSA encodes as a fixed spectral prior,
$C_{\mathrm{ex}}(t)=J_{F_{\mathrm{full}}}(h_{\mathrm{bg}})-L$ with $J_{F_{\mathrm{full}}}$ denoting the Jacobian of the full operator,
$r_{\mathrm{bg}}=F_{\mathrm{full}}(h_{\mathrm{bg}})-F_{\mathrm{bg}}(h_{\mathrm{bg}})$,
and $R_2$ contains second- and higher-order perturbation terms. The defect
$r_{\mathrm{bg}}$ is essential whenever the background does not satisfy the full PDE itself.
In implementation, we denote the structured background-dependent operator by $C_{\mathrm{bg}}$. For Burgers
and Klein--Gordon, $C_{\mathrm{bg}}=C_{\mathrm{ex}}$. For the heterogeneous wave equation,
$C_{\mathrm{bg}}$ uses a smoothed coefficient approximation, so the remaining coefficient mismatch is included in the unresolved perturbation term
\begin{equation}
R_{\mathrm{rem}}(t,\widetilde h)
=\bigl(C_{\mathrm{ex}}(t)-C_{\mathrm{bg}}(t)\bigr)\widetilde h
+R_2(t,\widetilde h).
\label{eq:remainder}
\end{equation}
For Burgers and Klein--Gordon, $C_{\mathrm{bg}}=C_{\mathrm{ex}}$ and hence
$R_{\mathrm{rem}}=R_2$.

\paragraph{Structured correction and neural closure.}
Define the spectral linear operator and background defect by
$\widehat L=\mathcal{T}L\mathcal{T}^{-1}$ and
$\widehat r_{\mathrm{bg}}=\mathcal{T}r_{\mathrm{bg}}$, respectively.
The implemented perturbation dynamics are
\begin{equation}
 \frac{d\widehat{\widetilde h}}{dt}
 =\widehat L\widehat{\widetilde h}
 +\mathcal{T}\!\left[C_{\mathrm{bg}}(t)\widetilde h\right]
 +\widehat r_{\mathrm{bg}}(t)
 +\varepsilon F_\theta(\widehat{\widetilde h},t).
 \label{eq:model}
\end{equation}
With $\varepsilon=0$, Eq.~\eqref{eq:model} is a deterministic nonautonomous linear correction at the target resolution. With $\varepsilon>0$, the closure approximates $R_{\mathrm{rem}}(t,\widetilde h)$ and any unresolved numerical effects.

\paragraph{Training loss.}
Let $\widehat G_\theta$ denote the right-hand side of Eq.~\eqref{eq:model}. Learned variants roll out the current perturbation trajectory and minimize a physical-space residual between the model right-hand side and the exact perturbation right-hand side:
\begin{align}
 \mathcal{L}_{\mathrm{pert}}
 =\sum_{i=1}^{N_t}\bigl\|
 &\mathcal{T}^{-1}\widehat G_\theta
 (t_i,\widehat{\widetilde h}_i) \nonumber\\
 &-F_{\mathrm{full}}(h_{\mathrm{bg},i}+\widetilde h_i)
 +F_{\mathrm{bg}}(h_{\mathrm{bg},i})
 \bigr\|_2^2,
 \label{eq:loss}
\end{align}
where $i=1,\ldots,N_t$ indexes sampled training times,
$h_{\mathrm{bg},i}=h_{\mathrm{bg}}(t_i)$,
$\widetilde h_i=\widetilde h(t_i)$, and the norm is the discrete physical-grid
$\ell_2$ norm over all physical-grid points and the relevant state variables at time $t_i$.

\paragraph{Three PDE instantiations.}
Table~\ref{tab:operators} summarizes the task-specific background operators and structured corrections.
For 2D Burgers with velocity $\boldsymbol q=(q_1,q_2)$ and viscosity $\nu$,
\begin{align}
 \partial_t\widetilde{\boldsymbol q}
 ={}&\nu\Delta\widetilde{\boldsymbol q}
 -(\boldsymbol q_{\mathrm{bg}}\!\cdot\nabla)\widetilde{\boldsymbol q}
 -(\widetilde{\boldsymbol q}\!\cdot\nabla)\boldsymbol q_{\mathrm{bg}}
 +r_{\mathrm{bg}}+R_2, \label{eq:burgers}
\end{align}
with $R_2=-(\widetilde{\boldsymbol q}\cdot\nabla)\widetilde{\boldsymbol q}$.

For Klein--Gordon with nonlinearity coefficient $m$, $u_t=v$ and $v_t=u_{xx}-m\sin u$. Choosing
$L(\widetilde u,\widetilde v)=(\widetilde v,\widetilde u_{xx}-m\widetilde u)$ gives
$C_{\mathrm{bg}}\widetilde h=(0,-m[\cos(u_{\mathrm{bg}})-1]\widetilde u)$.

For the heterogeneous wave equation with exact wave speed $c_{\mathrm{exact}}(y)$,
$v_t=c_{\mathrm{exact}}^2(y)\Delta u$. The background uses a smoothed profile
$c_{\mathrm{bg}}$. In the three-layer experiment,
$c_{\mathrm{exact}}(y)=c(y;1000)$ and $c_{\mathrm{bg}}(y)=c(y;s_{\mathrm{bg}})$, where
\[
\begin{aligned}
c(y;s)&=1+0.25\,\operatorname{sigmoid}(s(y-0.5))\\
&\quad+0.25\,\operatorname{sigmoid}(s(y-1.0)),\\
\operatorname{sigmoid}(z)&=(1+e^{-z})^{-1}.
\end{aligned}
\]
Using the unit-speed wave operator as the fixed part $L$, the implemented structured operator is
$C_{\mathrm{bg}}\widetilde h=(0,[c_{\mathrm{bg}}^2-1]\Delta\widetilde u)$,
while $r_{\mathrm{bg}}$ contains
$(0,[c_{\mathrm{exact}}^2-c_{\mathrm{bg}}^2]\Delta u_{\mathrm{bg}})$.
The residual target also includes the remaining coefficient-mismatch term
$(0,[c_{\mathrm{exact}}^2-c_{\mathrm{bg}}^2]\Delta\widetilde u)$ and numerical
interface effects.
Full derivations are provided in the supplement.

\begin{table*}[t]
\centering
\small
\begin{tabular}{llll}
\toprule
Task & Low-fidelity operator $F_{\mathrm{bg}}$ & Structured correction & Main unresolved effect \\
\midrule
2D Burgers & Truncated nonlinear solve
& Background-linearized advection & Quadratic advection \\
Klein--Gordon & Truncated nonlinear solve
& $-m[\cos(u_{\mathrm{bg}})-1]\widetilde u$ & Higher-order sine remainder \\
2D wave & Smoothed-coefficient wave solve
& $[c_{\mathrm{bg}}^2-1]\Delta\widetilde u$ & Coefficient/interface residual \\
\bottomrule
\end{tabular}
\caption{Task-specific background operators under the common perturbative decomposition. }
\label{tab:operators}
\end{table*}

\paragraph{Closure error trade-off.}
Equation~\eqref{eq:perturbation} makes the closure trade-off explicit. With $\varepsilon=0$, the structured solver leaves the local modeling error $R_{\mathrm{rem}}(t,\widetilde h)$.
Adding the closure changes the remaining error to
$R_{\mathrm{rem}}(t,\widetilde h)-\varepsilon F_\theta(\widehat{\widetilde h},t)$,
but also introduces estimation, optimization, and extrapolation errors. When the
perturbation is large, $R_{\mathrm{rem}}$ can provide a measurable closure-training signal, although
the first-order model is then less accurate. When the perturbation is small, the
closure target can fall below the scale of initialization and optimization noise.
Thus, adding $F_\theta$ need not improve the solver monotonically.

\paragraph{Implementation.}
Background trajectories are computed once with fourth-order Runge--Kutta
(RK4) integration and cached. Linear interpolation supplies background values
at RK4 substeps of the perturbation solver. Products with $C_{\mathrm{bg}}(t)$ are evaluated pseudo-spectrally:
coefficients are transformed to the physical grid, multiplied by the cached
background-dependent field, and projected back. For the Klein--Gordon and wave equations,
the neural closure modifies only $\partial_t\widetilde v$ because
$\partial_t\widetilde u=\widetilde v$ is exact. For Burgers, it modifies both
velocity derivatives.
Algorithm~\ref{alg:solver} summarizes background precomputation, optional closure training, and final reconstruction.

\paragraph{Pre-closure diagnostics.}
After the deterministic structured rollout and before closure training, we
evaluate the residual coefficients. Let
$\widehat{\widetilde h}_{\mathrm{str},i}$ denote the perturbation coefficients
produced by the structured-only rollout at time $t_i$. We define
\begin{equation}
 \widehat R_{\mathrm{str},i}
 =\widehat F_{\mathrm{pert}}^{\mathrm{ex}}
 (t_i,\widehat{\widetilde h}_{\mathrm{str},i})
 -\widehat G_0(t_i,\widehat{\widetilde h}_{\mathrm{str},i}),
 \label{eq:diagnostic-residual}
\end{equation}
where $\widehat F_{\mathrm{pert}}^{\mathrm{ex}}$ is the exact right-hand side
of Eq.~\eqref{eq:exact-perturbation} and $\widehat G_0$ is Eq.~\eqref{eq:model}
with $F_\theta\equiv0$. We use the shorthand
$\widehat F_{\mathrm{pert},i}^{\mathrm{ex}}=
\widehat F_{\mathrm{pert}}^{\mathrm{ex}}(t_i,\widehat{\widetilde h}_{\mathrm{str},i})$
and define the normalized residual magnitude by
$\rho_i=\|\widehat R_{\mathrm{str},i}\|_2/
\|\widehat F_{\mathrm{pert},i}^{\mathrm{ex}}\|_2$.
For the heterogeneous wave problem, we additionally measure the fraction of structured residual energy concentrated near coefficient interfaces. Let $R_{\mathrm{str}}^{\mathrm{phys}}(x,t_i)$ be the inverse transform
of the residual component entering $\partial_t\widetilde v$. For the three-layer wave profile, define the material-interface set by
$\Gamma=\{(x,y)\in\Omega:y=0.5\ \text{or}\ y=1.0\}$ and let
$\Omega_\delta=\{x\in\Omega:\operatorname{dist}(x,\Gamma)\le\delta\}$. The
interface residual fraction is
\begin{equation}
 \Lambda_\delta
 =
 \frac{\sum_i\int_{\Omega_\delta}
 |R_{\mathrm{str}}^{\mathrm{phys}}(x,t_i)|^2\,dx}
 {\sum_i\int_{\Omega}
 |R_{\mathrm{str}}^{\mathrm{phys}}(x,t_i)|^2\,dx}.
 \label{eq:interface-fraction}
\end{equation}
Here $\Omega$ is the physical domain, and $\Gamma$ contains the transition locations
of the wave-speed coefficient. Thus
$\Lambda_\delta$ is close to one when the residual energy is concentrated near
material interfaces and close to zero when little residual energy lies in that
interface band; $\Lambda_{0.1}$ uses a band of width $0.1$ around the
interfaces. We compute these diagnostics along the structured-only rollout and
aggregate them over the time window used for the corresponding closure study. These
quantities require one structured rollout but no closure optimization.

\subsection{Error View and Expected Regimes}

Let $\mathcal A(t)=L+C_{\mathrm{bg}}(t)$ and let $\Phi(t,\tau)$ denote its evolution
operator. Let $e_{\mathrm{str}}(t)$ be the error of the structured perturbation
relative to the exact perturbation around the same background. Variation of
constants gives the local error representation
\begin{equation}
 e_{\mathrm{str}}(t)
 \approx\int_0^t\Phi(t,\tau)R_{\mathrm{rem}}(\tau,\widetilde h(\tau))\,d\tau
 +e_{\mathrm{num}}(t),
 \label{eq:structured-error}
\end{equation}
where $e_{\mathrm{num}}$ collects time discretization, interpolation, and
spectral projection errors. The closure model replaces the integrand by
$R_{\mathrm{rem}}(\tau,\widetilde h(\tau))
-\varepsilon F_\theta(\widehat{\widetilde h}(\tau),\tau)$ and adds optimization
and generalization errors.
Equation~\eqref{eq:structured-error} separates the structured remainder,
numerical error, and learned-correction error, which is the organization used
in the regime experiments below.

This decomposition suggests three regimes. With a poor background,
$\widetilde h$ is large and the neglected remainder is measurable, but the
first-order approximation can itself be inaccurate. A neural closure may reduce
the error without eliminating the limits of the first-order model. At intermediate fidelity,
$\mathcal A(t)\widetilde h+r_{\mathrm{bg}}$ captures most of the correction, while a
learnable remainder may persist. For high-fidelity backgrounds in the Burgers and Klein--Gordon equations,
$R_{\mathrm{rem}}(t,\widetilde h)=R_2(t,\widetilde h)=\mathcal{O}(\|\widetilde h\|^2)$
for smooth nonlinearities. Since the structured terms have already removed the
leading-order perturbation response, the remaining closure target can be
comparable to optimization and initialization errors in the learned vector
field. In this regime, the learned closure may provide little benefit and can degrade extrapolation. The wave problem provides a useful counterpoint because
sharp-interface and discretization residuals can remain organized enough for
the chosen closure to learn, even when the structured error is small. Thus, closure
utility depends not only on residual magnitude, but also on how well the residual's spatial and
spectral structure matches the closure model.

\begin{algorithm}[t]
\caption{Perturbative Spectral Solver }
\label{alg:solver}
\begin{algorithmic}[1]
\REQUIRE initial state $h_0$, $F_{\mathrm{full}}$, $F_{\mathrm{bg}}$, basis, $\varepsilon$
\STATE Integrate Eq.~\eqref{eq:background} and cache $h_{\mathrm{bg}}(t_i)$.
\STATE Evaluate $r_{\mathrm{bg}}(t_i)$ at the cached times.
\STATE Initialize $\widetilde h_0=h_0-h_{\mathrm{bg}}(0)$.
\IF{$\varepsilon>0$}
\FOR{Adam optimization steps}
\STATE Roll out Eq.~\eqref{eq:model} over the training times with current $\theta$.
\STATE Evaluate the physical-space perturbation residual in Eq.~\eqref{eq:loss}.
\STATE Update $\theta$ with Adam.
\ENDFOR
\ENDIF
\STATE Roll out Eq.~\eqref{eq:model} at the target resolution with trained $\theta$,
or with $F_\theta\equiv0$ when $\varepsilon=0$.
\STATE Return $h(t_i)=h_{\mathrm{bg}}(t_i)+\widetilde h(t_i)$.
\end{algorithmic}
\end{algorithm}

%% file: src/experiments.tex
\section{Experiments}
\label{sec:experiments}

\paragraph{Protocol.}
We test our methods on 2D Burgers, 1D Klein--Gordon, and a heterogeneous 2D wave equation. For Burgers and Klein--Gordon, the background operator $F_{\mathrm{bg}}$ is a nonlinear spectral solver retaining $m_c$ modes per spatial dimension. For wave, $F_{\mathrm{bg}}$ uses a smoothed wave-speed coefficient. The perturbation is evolved at the target spectral resolution $M=201$, so $m_c<M$ for the truncated backgrounds. Table~\ref{tab:main} uses three seeds for P-NeuSA ($\varepsilon>0$); NeuSA uses seven seeds for the 2D tasks and three for Klein--Gordon. Background and P-NeuSA ($\varepsilon=0$) results are deterministic. All entries in Table~\ref{tab:main} use the relative discrete $\ell_2$ error
$E_{\mathrm{rel}}=\|\boldsymbol u_{\mathrm{pred}}-\boldsymbol u_{\mathrm{ref}}\|_{\ell_2}/\|\boldsymbol u_{\mathrm{ref}}\|_{\ell_2}$, where $\boldsymbol u$ stacks all reported field values at the sampled spatial points and output times in the indicated interval. For Burgers, we compute this error separately for $q_1$ and $q_2$ and average the two values. Thus, its two table rows differ only in their evaluation intervals. For closure studies, we take $G=E_{\mathrm{str}}/E_{\mathrm{cl}}$, where $E_{\mathrm{str}}$ and $E_{\mathrm{cl}}$ are the P-NeuSA ($\varepsilon=0$) and P-NeuSA ($\varepsilon>0$) errors, respectively. Thus, $G>1$ indicates that the neural closure improves the deterministic P-NeuSA variant. Further details on reference solvers, initial conditions, hyperparameters, complete tables, baselines, and visualizations are given in the supplement.

\paragraph{Task configurations.}
For 2D Burgers on $[0,4]^2$, we use periodic boundaries, $\nu=0.01$, a
nonlinear Fourier background with $m_c=51$, training on $t\in[0,1]$, and
evaluation through $t=2$. Klein--Gordon solves
$u_{tt}=u_{xx}-10\sin u$ on $[-4,4]$ with Gaussian displacement whose peak
amplitude is approximately $4$, zero initial velocity, and variable background
resolution $m_c$ over $t\in[0,3]$. The wave
experiment solves $u_{tt}=c_{\mathrm{exact}}^2(y)\Delta u$ on $[-2,2]^2$,
with $c_{\mathrm{exact}}(y)=c(y;1000)$ and a smoothed background
$c_{\mathrm{bg}}(y)=c(y;s_{\mathrm{bg}})$; the canonical setting uses
$s_{\mathrm{bg}}=50$ and $t\in[0,2]$. Initial conditions, basis choices, optimization
settings, and reference discretizations are detailed in the supplement.

\begin{table*}[t]
\centering
\small
\setlength{\tabcolsep}{2.5pt}
\begin{tabular}{lcccc}
\toprule
Task & Background &
\begin{tabular}{@{}c@{}}P-NeuSA\\($\varepsilon=0$)\end{tabular} &
\begin{tabular}{@{}c@{}}P-NeuSA\\($\varepsilon>0$)\end{tabular} & NeuSA \\
\midrule
2D Burgers, $[0,1]$
& $0.0666$ & $0.0025$ & $\mathbf{0.0019\pm0.0002}$ & $0.0600\pm0.0226$ \\
2D Burgers, $(1,2]$
& $0.0455$ & $\mathbf{0.0039}$ & $0.0186\pm0.0029$ & $0.1731\pm0.0302$ \\
Klein--Gordon, $m_c=101$, $[0,3]$
& $1.67\ee{-3}$ & $\mathbf{5.81\ee{-4}}$ & $6.35\ee{-4}\pm1.4\ee{-6}$ & $8.87\ee{-4}\pm4.7\ee{-5}$ \\
2D wave, $[0,2]$
& $0.0265$ & $0.0124$ & $\mathbf{0.010085}\pm3.3\ee{-5}$ & $0.0855\pm0.0306$ \\
\bottomrule
\end{tabular}
\caption{Canonical cross-task comparison using the windowed relative $\ell_2$
metric defined in the protocol. P-NeuSA ($\varepsilon=0$) is the deterministic
structured variant, while P-NeuSA ($\varepsilon>0$) includes the learned neural
closure. P-NeuSA ($\varepsilon>0$) entries report mean $\pm$ standard deviation
over three seeds. NeuSA uses seven seeds for the two 2D tasks and three for
Klein--Gordon; deterministic entries have no seed variation. Bold marks the lowest mean
error in each row. In P-NeuSA ($\varepsilon>0$), we use $\varepsilon=1.0$ for
Burgers, $0.5$ for Klein--Gordon, and $0.5$ for wave.}
\label{tab:main}
\end{table*}

\paragraph{Background correction.}
This experiment tests whether the structured perturbation itself improves a low-fidelity background before adding a neural closure.
Table~\ref{tab:main} shows that the correction reduces the background error by factors of $26.5$ on the Burgers equation, $2.9$ on the Klein--Gordon equation, and $2.1$ on the wave equation. The P-NeuSA errors are evaluated on the reconstructed field obtained by adding the high-resolution perturbation correction to the background trajectory.

\paragraph{Comparison with NeuSA.}
P-NeuSA ($\varepsilon=0$) is more accurate than NeuSA on all three tasks. For Burgers, it gives 24 times lower training error and 44 times lower extrapolation error. A 1000-step NeuSA control, rather than the canonical 200 steps, decreases these errors to $0.0145\pm0.0014$ and $0.0723\pm0.0080$, still 5.8 and 18.5 times above deterministic P-NeuSA. The Klein--Gordon and wave reductions are $34.5\%$ and $85.5\%$, respectively.

\paragraph{Burgers extrapolation.}
This test separates in-interval residual fitting from time extrapolation.
Although P-NeuSA ($\varepsilon>0$) slightly reduces the training-interval
relative $\ell_2$ error from $0.0025$ to $0.0019$, its extrapolation
error rises to $0.0186$. In contrast, P-NeuSA
($\varepsilon=0$) remains at $0.0039$, while NeuSA reaches $0.1731$. The
correction produced by the neural closure
therefore improves in-interval fitting but does not preserve this benefit
beyond the training horizon. Thus, a lower training residual does not
necessarily imply stable time extrapolation.

\begin{figure*}[!t]
    \centering
    \includegraphics[width=\textwidth]{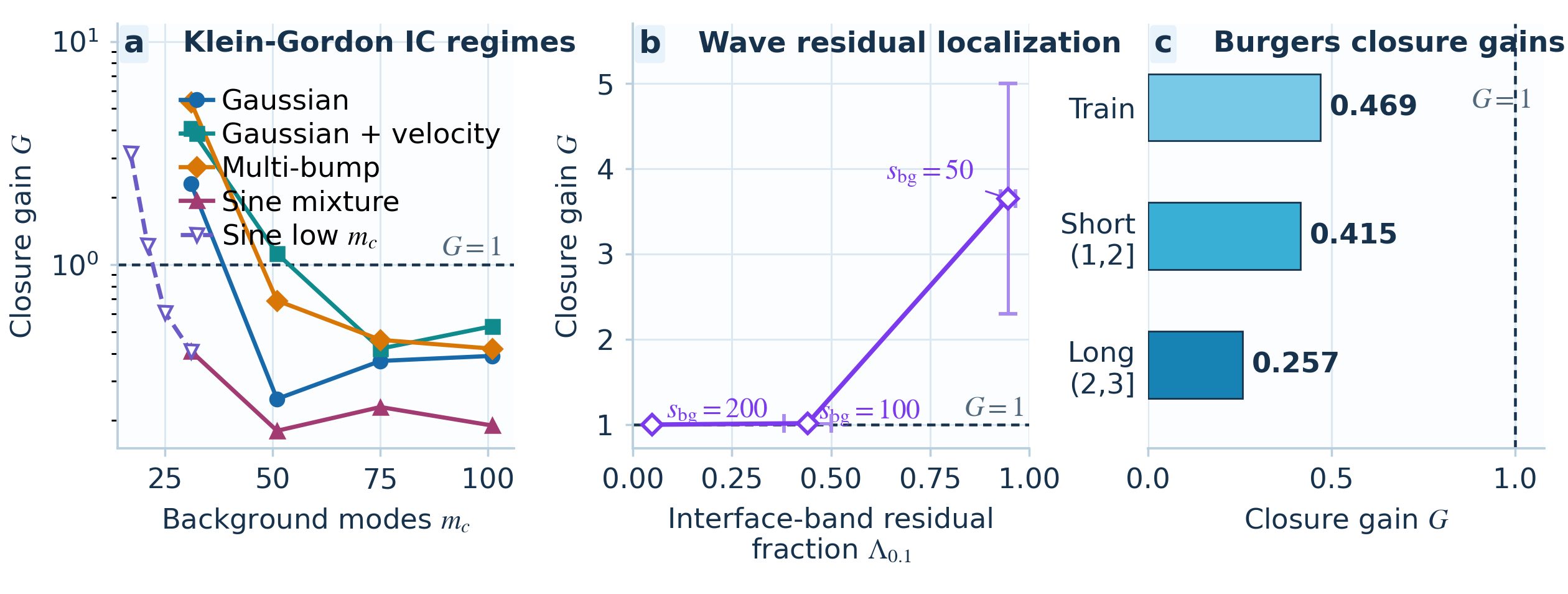}
    \caption{Operating-regime diagnostics for the optional neural closure. Gain is $G=E_{\mathrm{str}}/E_{\mathrm{cl}}$. For Klein--Gordon, candidate closure weights are screened in a finite preliminary sweep and the selected settings are confirmed with multiple seeds. Wave and Burgers use fixed task-specific weights, $\varepsilon=0.5$ and $1.0$, respectively. (a) Family-mean Klein--Gordon gains across background resolutions. The solid curves represent Gaussian displacement with zero or sinusoidal velocity, multi-bump displacement, and sine-mixture displacement; the dashed curve extends the sine-mixture study to $m_c=17,21,25$. (b) Wave mean gain versus the interface-band residual fraction $\Lambda_{0.1}$. Horizontal and vertical error bars denote standard deviations over four family-level values, each averaged over two initial conditions; the reported correlation uses all 24 initial-condition--background-sharpness pairs. (c) Burgers mean gains over 24 configurations.}
    \label{fig:main-results}
\end{figure*}

\paragraph{Klein--Gordon sweep protocol.}
We sweep Klein--Gordon background resolution
$m_c\in\{31,51,63,67,71,75,101\}$. At each resolution, the neural weight used for confirmation is selected from
$\varepsilon\in\{0.1,0.5,1.0\}$ in a single-seed selection sweep and then confirmed with three seeds. The resulting P-NeuSA ($\varepsilon>0$) entry is therefore the best-confirmed closure within this finite candidate set; the selected weights are listed in the Supplementary Material.

\paragraph{Regime-map results.}
At $m_c=31$, neural correction improves the structured result from $3.37\ee{-2}$ to $9.39\ee{-3}$ by a factor of $3.6$. The gain then decreases as the background resolution increases: the improvement factor is $1.14$ at $m_c=51$, is approximately one from $m_c=63$ to $71$, and drops below one at $m_c=75$ and $m_c=101$, where the closure slightly degrades the structured result. Nonlinear spectral broadening partly explains this trend: $m_c=51$ retains only $65.5\%$ of the energy in $\sin(u_{\mathrm{bg}})$, whereas $m_c=101$ retains $99.4\%$.

\paragraph{Regime-map interpretation.}
The structured correction reduces background error by factors of $2.9$--$33.8$
at all resolutions. Closure utility, however, declines from a clear benefit at
$m_c=31$, through changes comparable to seed variation at $m_c=63$--$71$, to
slight but consistent degradation at $m_c=101$.

\paragraph{Spectral origin of the transition.}
Although the initial state is well represented at $m_c=51$, its nonlinear
forcing is not. Let $a$ denote the amplitude of its dominant mode. For
$a\approx4$, direct projection shows that modes
$1$--$51$ and $1$--$101$ retain $65.5\%$ and $99.4\%$ of the energy in
$\sin(a\cos\varphi)$, respectively. Effective background resolution must
therefore capture the spectrum generated by the nonlinearity, not only that of
the initial condition. The complete harmonic analysis is provided in the
Supplementary Material.

\paragraph{Multi-initial-condition test.}
Figure~\ref{fig:main-results}(a) shows that the useful closure regime depends
on the initial-condition spectrum. At $m_c=31$, the closure helps the Gaussian
and multi-bump families but not the sine-mixture family. At $m_c=51$,
aggregate training/extrapolation gains are $0.56/0.49$, with only the Gaussian
family with sinusoidal velocity showing a local training benefit. For the
sine-mixture family, the transition occurs between $m_c=21$ and $25$.
Complete results are given in the Supplementary Material.

\paragraph{Residual organization.}
Figure~\ref{fig:main-results}(b) compares closure gain with interface
localization across eight initial conditions and
$s_{\mathrm{bg}}\in\{50,100,200\}$. Each of the 24 initial-condition and
background-sharpness pairs is one statistical unit, with closure errors averaged
over three neural seeds. As $s_{\mathrm{bg}}$ increases, mean
$\Lambda_{0.1}$ decreases from $0.948$ to $0.441$ and $0.049$, while mean $G$
decreases from $3.65$ to $1.02$ and approximately $1.00$. The normalized
residual magnitude does not track this trend. Across all 24 units,
$\Lambda_{0.1}$ correlates with $\log G$ (Pearson $r=0.847$,
$p=1.75\ee{-7}$). These results indicate that residual organization is more
informative about closure utility than residual magnitude alone. Family-level
results and additional statistics are reported under ``Wave Background
Fidelity and Residual Diagnostics'' in the Supplementary Material.

\paragraph{Burgers multi-initial-condition ablation.}
All 24 Burgers configurations have $G<1$.
Figure~\ref{fig:main-results}(c) reports mean training, short-extrapolation,
and long-extrapolation gains of $0.469$, $0.415$, and $0.257$, respectively.
The canonical result is consistent with this trend: the closure increases
extrapolation error from $0.0039$ to $0.0186$. Complete configurations and
per-case gains are given in the Supplementary Material.

\paragraph{Coordinate-network baselines.}
Among the physics-informed neural network (PINN), quadratic residual network
(QRes), first-layer sine network (FLS), and PINNsFormer baselines, QRes is the
strongest on all three tasks, but its errors remain 10--32 times larger than
those of P-NeuSA with $\varepsilon=0$. Complete training protocols, errors, and
times are reported in the Supplementary Material.

\paragraph{Computational cost.}
For P-NeuSA with $\varepsilon=0$, the complete deterministic evaluation takes
$0.59$ seconds for Klein--Gordon, $17.65$ seconds for 2D Burgers, and $5.95$
seconds for wave on two CPU threads. These totals include background
precomputation, the ODE rollout, and field reconstruction. NeuSA neural
optimization takes $716$, $329$, and $1513$ seconds, respectively, on one
NVIDIA A30. Because the CPU and GPU measurements use different hardware, we
report them as stage-specific wall-clock costs rather than normalized
speedups. The full breakdown is provided under ``Computational Cost'' in the
Supplementary Material.

%% file: src/discussion.tex
\section{Discussion and Limitations}
\label{sec:discussion}

\paragraph{Structured-correction regime.}
Perturbative-NeuSA is best interpreted as a defect-correction method in spectral coordinates. The method rewrites the target PDE around a background trajectory without changing the governing equation. The background supplies the reference trajectory, while Eq.~\eqref{eq:perturbation} corrects both its operator mismatch and its high-resolution response. This distinction explains why the final error can be substantially lower than the background error. The construction follows the coarse-model-plus-correction logic of multigrid and preconditioned iteration while evolving a time-dependent spectral correction.

\paragraph{Regimes for neural closure.}
The experiments show that neural closure has a conditional role rather than a
uniform benefit. When the background is too coarse, the residual can be large
enough to provide a training signal, but the first-order perturbation model may
also be less accurate. When the background is too accurate, the remaining
higher-order target can become too small to fit reliably. Between these cases,
the structured terms may already capture most of the useful correction. The
multi-initial-condition results show that this balance depends on the spectrum
generated by the initial state and the PDE, not on a fixed mode threshold. The
wave problem gives a case where closure remains useful: residuals localized near
interfaces are associated with closure gains even after the structured solver
has reduced the total error. The experiment on the Burgers equation gives the opposite case. Across the tested multi-initial-condition settings, the closure is not consistently beneficial
and substantially worsens the canonical extrapolation result.

\paragraph{Factors governing closure utility.}
The regime map should be read as a task-dependent diagnostic rather than a
universal rule based on mode count. The experiments identify three operational factors.
First, effective background fidelity depends on the spectrum generated by
both the PDE and the initial state. Second, residual organization is associated
with how well the unresolved error aligns with the closure model, as illustrated
by the wave interface diagnostic. Third, closure-model compatibility matters:
residual magnitude alone does not determine whether the tested MLP closure can fit
the remaining target reliably. The spectral-energy calculation for
$\sin(u)$ and the wave interface-energy diagnostic probe different mechanisms
behind this regime dependence, but neither provides a task-independent threshold.

\paragraph{Limitations.}
This study has five main limitations. First, it considers three smooth benchmark PDEs.
Shocks and strongly localized features may require different bases and
background solvers, or stabilization strategies. Second, the method assumes that a computationally inexpensive and
dynamically useful background operator is available. Third, we use MLP closures rather than architectures tailored to
small perturbation targets. Fourth, the closure weights in the Klein--Gordon regime map are
selected from a finite preliminary sweep and reported with three-seed
confirmation; adaptive weight selection is left for future work. Fifth, the
multi-initial-condition diagnostics use
designed families, and the Burgers ablation is representative rather than
exhaustive over closure architectures and optimization schedules. These
results therefore establish task-dependent trends, not universal thresholds.

\paragraph{Availability.}
Code, trained models, and reference data will be released upon publication.

%% file: src/conclusion.tex
\section{Conclusion}

Perturbative-NeuSA recasts neural spectral PDE solving as a perturbation problem
around a low-fidelity background trajectory. By separating the background solve,
the structured high-resolution correction, and the optional learned closure, the
method makes it possible to measure how much improvement comes from analytical
structure and how much comes from learning. Across the PDEs studied here, the
structured correction provides the dominant gain and can outperform NeuSA even
with $\varepsilon=0$, before any neural closure is trained.

The results support a conditional view of neural closure. Closure is useful when
the remaining residual is measurable and contains structure that the closure
model can represent. The multi-initial-condition studies show that this regime shifts with the
spectrum generated by the initial state and the PDE. The wave diagnostics show
an association between closure gains and interface-localized residuals, whereas the Burgers ablation
shows that the same closure can provide no net benefit and can worsen
extrapolation when the structured correction already captures the dominant
dynamics. Neural spectral solvers should therefore be evaluated using final
errors together with background fidelity, residual organization, and
closure-model compatibility.

%% file: src/supp_derivations.tex
\section{Complete Perturbation Derivations}

\subsection{General first-order system}

Let the target dynamics be $\partial_t h=F_{\mathrm{full}}(h)$ and the background satisfy
$\partial_t h_{\mathrm{bg}}=F_{\mathrm{bg}}(h_{\mathrm{bg}})$. With
$h=h_{\mathrm{bg}}+\widetilde h$,
\begin{align}
\partial_t\widetilde h
&=F_{\mathrm{full}}(h_{\mathrm{bg}}+\widetilde h)
  -F_{\mathrm{bg}}(h_{\mathrm{bg}}) \\
&=J_{F_{\mathrm{full}}}(h_{\mathrm{bg}})\widetilde h
  +r_{\mathrm{bg}}+R_2(t,\widetilde h),
\end{align}
where $J_{F_{\mathrm{full}}}(h_{\mathrm{bg}})
=D F_{\mathrm{full}}(h_{\mathrm{bg}})$ is the Jacobian of the target vector
field evaluated at the background and
$r_{\mathrm{bg}}=F_{\mathrm{full}}(h_{\mathrm{bg}})
-F_{\mathrm{bg}}(h_{\mathrm{bg}})$.
Splitting the Jacobian as
$J_{F_{\mathrm{full}}}(h_{\mathrm{bg}})=L+C_{\mathrm{ex}}(t)$ gives the
formulation used for Burgers and Klein--Gordon. The wave
experiment uses a smoothed-coefficient structured operator, as detailed below.

\subsection{2D Burgers}

For the velocity field $\boldsymbol q=(q_1,q_2)$,
\begin{equation}
\partial_t\boldsymbol q+(\boldsymbol q\cdot\nabla)\boldsymbol q
=\nu\Delta\boldsymbol q.
\end{equation}
Substitution of
$\boldsymbol q=\boldsymbol q_{\mathrm{bg}}+\widetilde{\boldsymbol q}$ gives
\begin{align}
\partial_t\widetilde{\boldsymbol q}
={}&\nu\Delta\widetilde{\boldsymbol q}
-(\boldsymbol q_{\mathrm{bg}}\cdot\nabla)\widetilde{\boldsymbol q}
-(\widetilde{\boldsymbol q}\cdot\nabla)\boldsymbol q_{\mathrm{bg}} \nonumber\\
&-(\widetilde{\boldsymbol q}\cdot\nabla)\widetilde{\boldsymbol q}
+r_{\mathrm{bg}},
\end{align}
with
\begin{equation}
r_{\mathrm{bg}}
=\nu\Delta\boldsymbol q_{\mathrm{bg}}
-(\boldsymbol q_{\mathrm{bg}}\cdot\nabla)\boldsymbol q_{\mathrm{bg}}
-\partial_t\boldsymbol q_{\mathrm{bg}}.
\end{equation}
The structured solver retains the first three perturbation terms and
$r_{\mathrm{bg}}$. The quadratic term is omitted when $\varepsilon=0$ and is part
of the learned target when $\varepsilon>0$.

\subsection{Klein--Gordon}

The equation
\begin{equation}
u_t=v,\qquad v_t=u_{xx}-m\sin u
\end{equation}
is decomposed as $u=u_{\mathrm{bg}}+\widetilde u$ and
$v=v_{\mathrm{bg}}+\widetilde v$. Taylor expansion gives
\begin{align}
\widetilde u_t&=\widetilde v+r_{\mathrm{bg},u},\\
\widetilde v_t&=\widetilde u_{xx}
-m\cos(u_{\mathrm{bg}})\widetilde u
+r_{\mathrm{bg},v}+R_2(t,\widetilde u).
\end{align}
For the consistent first-order background used here,
\begin{equation}
\begin{aligned}
r_{\mathrm{bg},u}
&=v_{\mathrm{bg}}-\partial_tu_{\mathrm{bg}}=0,\\
r_{\mathrm{bg},v}
&=u_{\mathrm{bg},xx}-m\sin(u_{\mathrm{bg}})
-\partial_tv_{\mathrm{bg}}.
\end{aligned}
\end{equation}
We use
\begin{align}
L(\widetilde u,\widetilde v)
&=(\widetilde v,\widetilde u_{xx}-m\widetilde u),\\
C_{\mathrm{bg}}(\widetilde u,\widetilde v)
&=(0,-m[\cos(u_{\mathrm{bg}})-1]\widetilde u).
\end{align}
The remainder is
\begin{equation}
R_2(t,\widetilde u)=-m\left[
\sin(u_{\mathrm{bg}}+\widetilde u)-\sin(u_{\mathrm{bg}})
-\cos(u_{\mathrm{bg}})\widetilde u
\right].
\end{equation}

\subsection{Heterogeneous 2D wave equation}

For
\begin{equation}
u_t=v,\qquad v_t=c_{\mathrm{exact}}^2(y)\Delta u,
\end{equation}
the background uses $c_{\mathrm{bg}}(y)$. Taking the homogeneous wave operator as
$L$ gives the exact Jacobian correction
\begin{align}
L(\widetilde u,\widetilde v)&=(\widetilde v,\Delta\widetilde u),\\
C_{\mathrm{ex}}(\widetilde u,\widetilde v)
&=(0,[c_{\mathrm{exact}}^2-1]\Delta\widetilde u).
\end{align}
The implementation uses the smoothed-coefficient structured operator
\begin{equation}
C_{\mathrm{bg}}(\widetilde u,\widetilde v)
=(0,[c_{\mathrm{bg}}^2-1]\Delta\widetilde u),
\end{equation}
and the background defect
\begin{equation}
r_{\mathrm{bg}}
=(0,[c_{\mathrm{exact}}^2-c_{\mathrm{bg}}^2]\Delta u_{\mathrm{bg}}).
\end{equation}
The target equation is linear in the state, so there is no Taylor remainder.
The residual target for the closure therefore contains the remaining
coefficient-mismatch perturbation
$(0,[c_{\mathrm{exact}}^2-c_{\mathrm{bg}}^2]\Delta\widetilde u)$ together with
discretization, interpolation, and unresolved sharp-interface effects.

%% file: src/supp_protocol.tex
\section{Experimental Protocol}

\paragraph{Common settings.}
The target spectral resolution is $M=201$ modes per spatial dimension and all
Neural ODE trajectories use fourth-order Runge--Kutta (RK4) integration. The
task-specific P-NeuSA closure architectures are summarized in
Table~\ref{tab:supp-architectures}. These closures use ReLU activations,
Xavier initialization~\cite{glorot2010understanding}, Adam
optimization~\cite{kingma2014adam}, learning-rate decay $\gamma=0.999$, and
gradient clipping with maximum norm $1.0$. Their canonical results use seeds
$42$, $43$, and $44$. NeuSA uses LeakyReLU, the default PyTorch linear-layer
initialization, Adam with $\gamma=0.999$, and no gradient clipping. Its results
use seeds $42$--$48$ for the two 2D tasks and $42$--$44$ for Klein--Gordon.

\paragraph{2D Burgers.}
The domain is $[0,4]^2$ with periodic boundaries, $\nu=0.01$,
$q_{1,0}=\sin(\pi x)\sin(\pi y)$, and
$q_{2,0}=\cos(\pi x)\cos(\pi y)$. Training covers $[0,1]$ and extrapolation covers
$(1,2]$. The nonlinear background uses $m_c=51$ modes per dimension.
P-NeuSA ($\varepsilon>0$) uses 500 optimization steps, learning rate
$10^{-3}$, and $\varepsilon=1.0$. The canonical NeuSA baseline uses 200 steps,
learning rate $5\ee{-3}$, and model weight $0.1$. An additional NeuSA control
uses 1000 steps with otherwise identical settings
to test whether a longer training budget narrows the error gap to P-NeuSA
($\varepsilon=0$).

\paragraph{Klein--Gordon.}
The domain is $[-4,4]$ with zero Dirichlet boundaries, $m=10$, and sine basis.
The initial condition is
$u_0=(2\pi\sigma^2)^{-1/2}\exp[-x^2/(2\sigma^2)]$ with $\sigma=0.1$ and
$v_0=0$. The canonical trajectory and regime map use $t\in[0,3]$. P-NeuSA
uses 1000 steps and learning rate $10^{-2}$, whereas NeuSA uses 1000 steps,
learning rate $2\ee{-2}$, and model weight $0.1$. The regime
map considers $m_c\in\{31,51,63,67,71,75,101\}$ and
$\varepsilon\in\{0.1,0.5,1.0\}$. The canonical $m_c=101$ configuration uses
$\varepsilon=0.5$.

\paragraph{2D wave.}
The physical evaluation domain is $[-2,2]^2$. The cosine representation is
defined on $[-4,4]^2$ to reduce boundary artifacts, and the temporal domain is
$t\in[0,2]$. The Gaussian initial
condition has $\sigma=0.1$ and zero initial velocity. The exact wave-speed
profile uses sigmoid sharpness $s=1000$. The background uses $s=50$.
NeuSA uses 2000 steps, learning rate $10^{-2}$, and model weight $1.0$.
The perturbative model uses 1000 steps, learning rate $5\ee{-3}$, and
$\varepsilon=0.5$.

\paragraph{Multi-initial-condition diagnostics.}
The Klein--Gordon study trains on $t\in[0,3]$ and evaluates extrapolation on
$t\in(3,5]$. It uses four families: Gaussian displacement with zero
velocity, Gaussian displacement with sinusoidal velocity, multi-bump
displacement, and sine-mixture displacement. Each family contains three
initial conditions. Their parameterized formulas are given below under
``Initial-Condition Families.''
The main grid uses $m_c\in\{31,75,101\}$ with
$\varepsilon\in\{0.1,1.0,0.5\}$, respectively, and closure seeds
$\{0,1,2,3,4\}$. An additional $m_c=51,\varepsilon=0.1$ run uses the same
12 initial conditions and seeds. The targeted sine-mixture study uses
$m_c\in\{5,9,13,17,21,25\}$ and $\varepsilon=0.1$; the final
operating-regime analysis reports the multi-seed confirmation at
$m_c\in\{17,21,25\}$ with seeds $\{0,1,2,3,4\}$.

The wave study uses $t\in[0,2]$ and centered Gaussian, off-center Gaussian, multi-pulse, and
cosine-field initial conditions, with two instances per family, background
sharpness $s_{\mathrm{bg}}\in\{50,100,200\}$, fixed task-specific weight
$\varepsilon=0.5$, and closure seeds $\{0,1,2\}$.

The Burgers multi-initial-condition ablation uses $t\in[0,3]$, with training on
$[0,1]$, short extrapolation on $(1,2]$, and long extrapolation on $(2,3]$.
It uses one representative initial
condition from each of the low-Fourier, medium-Fourier, and vortex-like
families, viscosities $\nu\in\{0.01,0.02\}$,
$m_c\in\{31,51,75,101\}$, fixed task-specific weight $\varepsilon=1.0$, three
closure seeds $\{0,1,2\}$, and 500 optimization steps.
Residual diagnostics are computed from the deterministic structured rollout
before any closure training.

For the two-component Burgers field $\boldsymbol q=(q_1,q_2)$, errors over a
sample set $\mathcal W$ use the componentwise average
\begin{equation}
E_{\mathrm{avg}}(\mathcal W)
=\frac{1}{2}\sum_{j=1}^{2}
\frac{\|q_{j,\mathrm{pred}}-q_{j,\mathrm{ref}}\|_{\ell_2(\mathcal W)}}
{\|q_{j,\mathrm{ref}}\|_{\ell_2(\mathcal W)}}.
\end{equation}
The canonical main-text comparison and Table~\ref{tab:supp-burgers} use
$E_{\mathrm{avg}}(\mathcal W)$ directly,
stacking every spatial point and output time in the indicated window
$\mathcal W$. Thus, $q_1$ and $q_2$ receive equal weight independently of
their reference energies. The multi-initial-condition extrapolation diagnostic
instead evaluates the same componentwise average at each output time before
averaging over the short and long extrapolation windows.

\paragraph{Reference solutions.}
Klein--Gordon uses a pseudo-spectral RK4 solution~\cite{boyd2001chebyshev} with
$\Delta x=0.04$ and $\Delta t=0.001$, evaluated at 201 spatial points and 201
output times.
Burgers uses pseudo-spectral RK4 on $t\in[0,2]$ with
$\Delta x=\Delta y=0.02$ and internal step $\Delta t=0.001$, evaluated on a
$201\times201$ spatial grid at 401 output times separated by $0.005$. Wave reference data use eighth-order finite
differences in space and second-order time integration with
$\Delta x=\Delta y=0.01$ and $\Delta t=0.001$, evaluated on a
$101\times101$ spatial grid at 201 output times.

%% file: src/supp_results.tex
\section{Full Numerical Results}

The first tables report the canonical task-level comparisons; the subsequent
tables report the operating-regime diagnostics.

\begin{table*}[t]
\centering
\begin{tabular}{lccc}
\toprule
Method & Relative $\ell_2$ $[0,1]$ & Relative $\ell_2$ $(1,2]$ & Relative $\ell_2$ $[0,2]$ \\
\midrule
Background & $0.0666$ & $0.0455$ & $0.0640$ \\
P-NeuSA ($\varepsilon=0$) & $0.0025$ & $0.0039$ & $0.0028$ \\
P-NeuSA ($\varepsilon>0$) & $0.0019\pm0.0002$ & $0.0186\pm0.0029$ & $0.0073\pm0.0011$ \\
NeuSA & $0.0600\pm0.0226$ & $0.1731\pm0.0302$ & $0.0866\pm0.0222$ \\
\bottomrule
\end{tabular}
\caption{Canonical 2D Burgers results. Each column reports the component-averaged
relative $\ell_2$ error after stacking all spatial points and output times in
the indicated interval.
The P-NeuSA ($\varepsilon>0$) row reports three seeds, and NeuSA reports seven.}
\label{tab:supp-burgers}
\end{table*}

\begin{table*}[t]
\centering
\begin{tabular}{ccccc}
\toprule
$m_c$ & Background & P-NeuSA ($\varepsilon=0$) & P-NeuSA ($\varepsilon>0$) & Ratio \\
\midrule
31 & $2.92\ee{-1}$ & $3.37\ee{-2}$ & $9.39\ee{-3}\pm2.6\ee{-3}$ & $3.60$ \\
51 & $6.60\ee{-2}$ & $1.95\ee{-3}$ & $1.71\ee{-3}\pm8.4\ee{-5}$ & $1.14$ \\
63 & $2.30\ee{-2}$ & $1.15\ee{-3}$ & $1.15\ee{-3}\pm1.3\ee{-7}$ & $1.00$ \\
67 & $1.61\ee{-2}$ & $1.07\ee{-3}$ & $1.09\ee{-3}\pm3.5\ee{-5}$ & $0.98$ \\
71 & $1.15\ee{-2}$ & $1.00\ee{-3}$ & $1.00\ee{-3}\pm7.7\ee{-8}$ & $1.00$ \\
75 & $8.32\ee{-3}$ & $9.27\ee{-4}$ & $9.62\ee{-4}\pm4.6\ee{-7}$ & $0.96$ \\
101 & $1.67\ee{-3}$ & $5.81\ee{-4}$ & $6.35\ee{-4}\pm1.4\ee{-6}$ & $0.91$ \\
\bottomrule
\end{tabular}
\caption{Klein--Gordon regime map. The P-NeuSA ($\varepsilon>0$) column reports
the best-confirmed closure within the finite $\varepsilon$ sweep. Ratio is
P-NeuSA ($\varepsilon=0$) relative $\ell_2$ error divided by selected P-NeuSA
($\varepsilon>0$) relative $\ell_2$ error. Values above one indicate that the
neural closure improves on the deterministic variant.}
\label{tab:supp-regime}
\end{table*}

\begin{table}[t]
\centering
\small
\setlength{\tabcolsep}{2pt}
\begin{tabular}{cccccccc}
\toprule
$m_c$ & 31 & 51 & 63 & 67 & 71 & 75 & 101 \\
\midrule
Selected $\varepsilon$ & 0.1 & 0.1 & 1.0 & 0.1 & 1.0 & 1.0 & 0.5 \\
\bottomrule
\end{tabular}
\caption{Closure weights used in the confirmed Klein--Gordon regime
map. At each background resolution, the weight is selected from
$\{0.1,0.5,1.0\}$ by the preliminary sweep before the reported three-seed
evaluation.}
\label{tab:supp-kg-closure-weights}
\end{table}

\begin{table}[t]
\centering
\small
\setlength{\tabcolsep}{2pt}
\begin{tabular}{lcc}
\toprule
Method & Relative $\ell_2$ & Opt. time (s) \\
\midrule
NeuSA & $0.0855\pm0.0306$ & $1513\pm117$ \\
Background & $0.0265$ & --- \\
P-NeuSA ($\varepsilon=0$) & $0.0124$ & --- \\
P-NeuSA ($\varepsilon=0.1$) & $0.010106\pm6.2\ee{-5}$ & $3010$ \\
P-NeuSA ($\varepsilon=0.5$) & $0.010085\pm3.3\ee{-5}$ & $3059$ \\
\bottomrule
\end{tabular}
\caption{Canonical heterogeneous 2D wave comparison. P-NeuSA neural
rows use three seeds and NeuSA uses seven. Optimization time is measured on one
NVIDIA A30.}
\label{tab:supp-wave}
\end{table}

\begin{table}[t]
\centering
\begin{tabular}{lccc}
\toprule
Burgers setting & Train & Extrap. $(1,2]$ & Long $(2,3]$ \\
\midrule
Low Fourier & $0.542$ & $0.533$ & $0.366$ \\
Medium Fourier & $0.330$ & $0.370$ & $0.242$ \\
Vortex-like & $0.534$ & $0.341$ & $0.163$ \\
\midrule
Overall & $0.469$ & $0.415$ & $0.257$ \\
\bottomrule
\end{tabular}
\caption{Mean Burgers gain $G$ over 24 configurations in the 500-step
multi-initial-condition ablation: three initial conditions, two viscosities,
and four background resolutions, each with three closure seeds. Errors use the
componentwise average defined in the experimental protocol.}
\label{tab:supp-burgers-multi-ic}
\end{table}

\begin{table}[t]
\centering
\small
\setlength{\tabcolsep}{2pt}
\begin{tabular}{@{}lcccc@{}}
\toprule
Initial-condition family & $m_c=31$ & $m_c=51$ & $m_c=75$ & $m_c=101$ \\
\midrule
Gaussian, zero $v$ & $2.31$ & $0.25$ & $0.37$ & $0.39$ \\
Gaussian, sinusoidal $v$ & $4.08$ & $1.12$ & $0.42$ & $0.53$ \\
Multi-bump & $5.35$ & $0.69$ & $0.46$ & $0.42$ \\
Sine mixture & $0.41$ & $0.18$ & $0.23$ & $0.19$ \\
\bottomrule
\end{tabular}
\caption{Mean Klein--Gordon training gain $G$ over three initial
conditions and five closure seeds per family. The $m_c=51$ column pairs each
closure run with the corresponding deterministic P-NeuSA baseline.}
\label{tab:supp-kg-multi-ic}
\end{table}

\begin{table}[t]
\centering
\small
\setlength{\tabcolsep}{2pt}
\begin{tabular}{@{}lcccc@{}}
\toprule
Initial-condition family & $m_c=31$ & $m_c=51$ & $m_c=75$ & $m_c=101$ \\
\midrule
Gaussian, zero $v$ & $1.42$ & $0.24$ & $0.48$ & $0.38$ \\
Gaussian, sinusoidal $v$ & $2.21$ & $0.90$ & $0.62$ & $0.60$ \\
Multi-bump & $2.22$ & $0.65$ & $0.56$ & $0.40$ \\
Sine mixture & $0.35$ & $0.16$ & $0.32$ & $0.25$ \\
\bottomrule
\end{tabular}
\caption{Mean Klein--Gordon extrapolation gain $G$ on $t\in(3,5]$
over three initial conditions and five closure seeds per family. The $m_c=51$
column pairs each closure run with the corresponding deterministic P-NeuSA
baseline.}
\label{tab:supp-kg-multi-ic-extrap}
\end{table}

\begin{table}[t]
\centering
\begin{tabular}{ccc}
\toprule
$m_c$ & Training gain & Extrapolation gain \\
\midrule
17 & $3.16$ & $1.61$ \\
21 & $1.22$ & $1.35$ \\
25 & $0.61$ & $0.59$ \\
\bottomrule
\end{tabular}
\caption{Targeted sine-mixture low-resolution study, averaged over
three initial conditions with multi-seed closure confirmation for
$m_c=17,21,25$. Training gain uses $t\in[0,3]$ and extrapolation gain uses
$t\in(3,5]$.}
\label{tab:supp-kg-lowmc}
\end{table}

\section{Nonlinear Spectral Broadening}

For an approximately single-frequency state with amplitude $a$,
\begin{equation}
\sin(a\cos\varphi)
=2\sum_{n=0}^{\infty}(-1)^nJ_{2n+1}(a)
\cos((2n+1)\varphi).
\end{equation}
Here $J_k$ is the Bessel function of the first kind.
At a Klein--Gordon peak amplitude of approximately $4$, the nonlinear term occupies
substantially more modes than the initial condition. Direct spectral analysis
shows that modes $1$--$51$ contain $65.5\%$ of the energy in $\sin(u_0)$,
modes $1$--$75$ contain $97.5\%$, and modes $1$--$101$ contain $99.4\%$.
By comparison, modes $1$--$51$ already contain $99.6\%$ of the energy in
$u_0$, and modes $1$--$101$ contain effectively all of it. Thus, $m_c=51$ can
resolve the initial state while omitting a substantial part of the spectrum
generated by the nonlinear forcing. The background resolution must therefore
be selected from the spectral support of the nonlinear term, not from the
initial condition alone. This contrast explains the large background defect at
$m_c=51$ and its rapid reduction by $m_c=101$.

\clearpage

\input{src/supp_initial_conditions}

\section{Multi-Initial-Condition Diagnostics}

We define closure gain as $G=E_{\mathrm{str}}/E_{\mathrm{cl}}$, so values above
one indicate improvement over P-NeuSA ($\varepsilon=0$). Tables~\ref{tab:supp-kg-multi-ic}
and~\ref{tab:supp-kg-multi-ic-extrap} show that the absolute background
resolution does not determine closure utility across Klein--Gordon initial
conditions. At $m_c=31$, the closure helps three families but degrades the
sine-mixture family. At $m_c=51$, the aggregate training and extrapolation
gains over all 12 initial conditions are $0.56$ and $0.49$. The only family
with a training gain above one is Gaussian displacement with sinusoidal
velocity, and its extrapolation gain remains below one. At $m_c=75$ and $101$,
all family means are below one.

All 24 Burgers configurations have $G<1$. The closest case is the low-Fourier
initial condition with $\nu=0.02$ and $m_c=31$, whose training, short
extrapolation, and long extrapolation gains are $0.953$, $0.954$, and $0.894$.

\input{src/supp_wave_background}

\paragraph{Additional solution visualizations.}
Figure~\ref{fig:supp-burgers} shows 2D Burgers;
Figure~\ref{fig:supp-kg}, Klein--Gordon; and
Figure~\ref{fig:supp-wave}, the heterogeneous wave equation.

\section{Coordinate-Network Baselines}
We evaluate the physics-informed neural network (PINN), quadratic residual
network (QRes), first-layer sine network (FLS), and PINNsFormer baselines used
by NeuSA. Klein--Gordon runs use 10,000 Adam steps; Burgers and wave use 20,000
steps. The learning rate is $10^{-3}$, and each entry averages three seeds.

\begin{table*}[!t]
\centering
\begin{tabular}{llcc}
\toprule
Equation & Model & Relative $\ell_2$ & Training time (s) \\
\midrule
Klein--Gordon & PINN & $1.32\ee{-1}\pm3.06\ee{-2}$ & $2389\pm39$ \\
 & QRes & $1.84\ee{-2}\pm9.26\ee{-3}$ & $2364\pm22$ \\
 & FLS & $1.24\ee{-1}\pm4.73\ee{-2}$ & $2430\pm43$ \\
 & PINNsFormer & $7.71\ee{-1}\pm2.37\ee{-1}$ & $5899\pm27$ \\
\midrule
2D Burgers & PINN & $1.82\ee{-1}\pm5.29\ee{-2}$ & $2734\pm33$ \\
 & QRes & $7.10\ee{-2}\pm1.57\ee{-3}$ & $2884\pm22$ \\
 & FLS & $1.97\ee{-1}\pm5.06\ee{-2}$ & $2834\pm15$ \\
 & PINNsFormer & $1.05\pm1.10\ee{-1}$ & $5821\pm84$ \\
\midrule
2D wave & PINN & $5.74\ee{-1}\pm1.79\ee{-1}$ & $1840\pm7$ \\
 & QRes & $1.28\ee{-1}\pm3.15\ee{-2}$ & $1903\pm23$ \\
 & FLS & $6.46\ee{-1}\pm1.98\ee{-1}$ & $1856\pm33$ \\
 & PINNsFormer & $1.06\pm1.63\ee{-1}$ & $3830\pm29$ \\
\bottomrule
\end{tabular}
\caption{PINN-family baselines. Times are measured on one NVIDIA A30.}
\label{tab:supp-baselines}
\end{table*}

\section{Computational Cost}
Table~\ref{tab:supp-cost} reports the wall-clock breakdown for the canonical NeuSA and P-NeuSA
configurations.

\begin{table*}[!t]
\centering
\small
\setlength{\tabcolsep}{2pt}
\begin{tabular}{@{}llrrrrr@{}}
\toprule
Equation & Method & Background precomp. & Neural optimization & ODE rollout & Reconstruction & Stage sum \\
\midrule
Klein--Gordon & NeuSA & --- & $715.6\pm1.4$ & $0.351\pm0.008$ & --- & $716.0\pm1.4$ \\
 & P-NeuSA ($\varepsilon=0$) & $0.10$ & --- & $0.49$ & $0.001$ & $0.59$ \\
 & P-NeuSA ($\varepsilon>0$) & $0.10$ & $1700$ & $1.27$ & --- & $1701.37$ \\
\midrule
2D Burgers & NeuSA & --- & $328.9\pm23.2$ & $0.690\pm0.020$ & --- & $329.6\pm23.2$ \\
 & P-NeuSA ($\varepsilon=0$) & $4.22$ & --- & $13.00$ & $0.420$ & $17.65$ \\
 & P-NeuSA ($\varepsilon>0$) & $4.22$ & $4662$ & $6.50$ & --- & $4672.72$ \\
\midrule
2D wave & NeuSA & --- & $1513\pm117$ & $0.378\pm0.040$ & --- & $1514\pm117$ \\
 & P-NeuSA ($\varepsilon=0$) & $1.05$ & --- & $4.38$ & $0.510$ & $5.95$ \\
 & P-NeuSA ($\varepsilon>0$) & $1.05$ & $3059$ & $2.28$ & --- & $3062.33$ \\
\bottomrule
\end{tabular}
\caption{Wall-clock seconds. Deterministic timings are medians of three runs
after one warm-up on two Intel Xeon Gold 6226R CPU threads. They reproduce the
canonical relative $\ell_2$ errors for all three tasks. Background timings for
P-NeuSA ($\varepsilon>0$) use the same independently timed background operator.
Neural optimization and neural-model rollout are measured on one NVIDIA A30.
NeuSA uses seven seeds for the two 2D tasks and three for Klein--Gordon;
P-NeuSA uses three seeds. Stage sum is the sum of the measured entries in
that row. Reconstruction was not separately instrumented for neural methods
and is therefore not included in their stage sums.}
\label{tab:supp-cost}
\end{table*}

\begin{table*}[!t]
\centering
\small
\setlength{\tabcolsep}{3pt}
\begin{tabular}{@{}p{0.16\textwidth}p{0.22\textwidth}p{0.36\textwidth}p{0.20\textwidth}@{}}
\toprule
Model & Conditioning and input & Hidden architecture & Predicted output \\
\midrule
Klein--Gordon closure
& Time and both perturbation-state coefficient vectors
& Two hidden layers of width $4M=804$, with bias
& $\partial_t\widetilde v$ coefficients \\
Burgers closure
& Time and the coupled two-component perturbation state
& Two transpose-network branches, widening factor $2$, no additional hidden blocks, no bias
& Both $\partial_t\widetilde q_j$ fields \\
Wave closure
& Displacement perturbation coefficients; no explicit time input
& One row/column transpose network, widening factor $2$, no additional hidden blocks, no bias
& $\partial_t\widetilde v$ coefficients \\
PINN
& Space--time coordinates
& Four width-$512$ hidden layers with $\tanh$ activations
& Physical field values \\
QRes
& Space--time coordinates
& Width $256$; one input quadratic block and four quadratic residual blocks
& Physical field values \\
FLS
& Space--time coordinates
& Four width-$512$ hidden layers; sine activation in the first layer and $\tanh$ thereafter
& Physical field values \\
PINNsFormer
& Space--time coordinate sequences
& Model dimension $32$, one encoder and decoder layer, two attention heads, feed-forward width $256$, and two width-$512$ output layers
& Physical field values \\
\bottomrule
\end{tabular}
\caption{Neural architectures used in the reported experiments. Here $M=201$
is the target spectral resolution. The Burgers branches share the coupled state
input but predict the two velocity derivatives separately. The wave closure is
state-dependent but not explicitly time-conditioned.}
\label{tab:supp-architectures}
\end{table*}

%% file: src/supp_initial_conditions.tex
\section{Initial-Condition Families}
\enlargethispage{4.2in}

\paragraph{Klein--Gordon.}
Let $\psi_k(x)=\sin[k\pi(x+4)/8]$ and
$g(A,\mu,\sigma;x)=A\exp[-(x-\mu)^2/(2\sigma^2)]$.
The Gaussian-displacement families use either zero velocity or a sinusoidal
velocity in the first sine mode:
\begin{align}
(u_0,v_0)&=(g(A,\mu,\sigma;x),0),\\
(u_0,v_0)&=(g(A,\mu,\sigma;x),\alpha\psi_1(x)).
\end{align}
The multi-bump family superposes three Gaussian displacements and uses zero
velocity:
\begin{equation}
(u_0,v_0)=\left(\sum_{r=1}^{3}g(A_r,\mu_r,\sigma_r;x),0\right).
\end{equation}
The sine-mixture family represents both displacement and velocity by eight
sine modes:
\begin{equation}
(u_0,v_0)=\left(\sum_{k=1}^{8}a_k\psi_k(x),
                       \sum_{k=1}^{8}b_k\psi_k(x)\right).
\end{equation}
For the first two families, $A\in[2,5]$, $\mu\in[-1.5,1.5]$, and
$\sigma\in[0.08,0.25]$, with $\alpha\in[-0.5,0.5]$ for the second. For the
multi-bump family, $A_r\in[0.8,2.5]$, $\mu_r\in[-2.5,2.5]$, and
$\sigma_r\in[0.08,0.20]$. The sine-mixture coefficients have standard
deviation $k^{-2}$ and are jointly rescaled so that
$\max_x|u_0(x)|\in[2,5]$.

\paragraph{Heterogeneous wave equation.}
The centered- and off-center-Gaussian families share
\begin{equation}
u_0^{\mathrm{G}}
 =A\exp\!\left[-\frac{(x-x_0)^2+(y-y_0)^2}{2\sigma^2}\right].
\end{equation}
The centered family fixes $(x_0,y_0)=(0,0)$, whereas the off-center family
samples $(x_0,y_0)\in[-1,1]^2$. The multi-pulse family superposes three
Gaussian pulses:
\begin{equation}
u_0^{\mathrm{MP}}
 =\sum_{r=1}^{3}A_r
\exp\!\left[-\frac{(x-x_r)^2+(y-y_r)^2}{2\sigma_r^2}\right].
\end{equation}
Let
$\mathcal K=\{(k_x,k_y):0\leq k_x,k_y\leq5,\ (k_x,k_y)\neq(0,0)\}$.
The cosine-field family combines the modes in $\mathcal K$ with random phases:
\begin{equation}
u_0
=\sum_{\boldsymbol k\in\mathcal K}a_{\boldsymbol k}
\cos\!\left(\tfrac{\pi}{8}[k_x(x+4)+k_y(y+4)]+\theta_{\boldsymbol k}\right).
\end{equation}
All families use $v_0=0$. Gaussian amplitudes lie in $[0.8,1.2]$, widths in
$[0.18,0.35]$, and off-center locations in $[-1,1]^2$; multi-pulse amplitudes
lie in $[0.3,0.8]$, widths in $[0.15,0.35]$, and locations in
$[-1.5,1.5]^2$. Cosine coefficients have standard deviation
$(1+k_x^2+k_y^2)^{-1}$, random phases, and are rescaled to unit peak
amplitude.

\paragraph{2D Burgers.}
Let $\phi_{\boldsymbol k}
=\frac{\pi}{2}(k_xx+k_yy)+\theta_{\boldsymbol k}$.
The low- and medium-Fourier families use the same random Fourier form,
\begin{equation}
\boldsymbol q_0
=\left(\sum_{\boldsymbol k}a_{\boldsymbol k}\sin\phi_{\boldsymbol k},
        \sum_{\boldsymbol k}b_{\boldsymbol k}\cos\phi_{\boldsymbol k}\right),
\end{equation}
\pagebreak[4]
but retain modes with $|k_x|,|k_y|\leq3$ and $5$, respectively. Their
coefficients have standard deviation $(1+k_x^2+k_y^2)^{-2}$. The vortex-like
family constructs a divergence-free velocity field from a streamfunction,
$\boldsymbol q_0=(\partial_y\psi,-\partial_x\psi)$, where $\psi$ is a random
sine series over $k_x^2+k_y^2\leq16$. Each field is normalized so that
$\operatorname{mean}(q_{1,0}^2+q_{2,0}^2)=0.5$.

%% file: src/supp_wave_background.tex
\section{Wave Background Fidelity and Residual Diagnostics}

\paragraph{Background construction.}
The framework requires a low-fidelity operator, not a single universal type of
approximation. Burgers and Klein--Gordon are nonlinear, and their principal
background error is unresolved nonlinear spectral broadening. A truncated
nonlinear solve therefore tests the intended correction mechanism.

The wave benchmark is linear in the state. If the background used the exact
sharp coefficient and only fewer modes, the closure target would be dominated
by projection error rather than coefficient mismatch. We instead smooth the
material interfaces from sharpness $s=1000$ to $s=50$. The perturbation then
corrects a coefficient mismatch between the smoothed background profile and
the sharp target wave-speed profile while retaining the same decomposition:
\begin{equation}
r_{\mathrm{bg}}
=(0,[c_{\mathrm{exact}}^2-c_{\mathrm{bg}}^2]\Delta u_{\mathrm{bg}}).
\end{equation}

\paragraph{Residual diagnostics.}
We vary $s_{\mathrm{bg}}\in\{50,100,200\}$ to examine how background fidelity
changes the residual presented to the neural closure. We use the closure gain
$G=E_{\mathrm{str}}/E_{\mathrm{cl}}$, the interface residual fraction
$\Lambda_{0.1}$, and the normalized residual magnitude
$\rho_i=\|\widehat R_{\mathrm{str},i}\|_2/
\|\widehat F_{\mathrm{pert},i}^{\mathrm{ex}}\|_2$.
Here $\Lambda_{0.1}$ is the fraction of physical-space residual energy within
distance $0.1$ of the wave-speed interfaces.

The mean gain across eight initial conditions decreases from $3.65$ at
$s_{\mathrm{bg}}=50$ to $1.02$ at $s_{\mathrm{bg}}=100$ and approximately
$1.00$ at $s_{\mathrm{bg}}=200$. The corresponding mean fractions of residual
energy within distance $0.1$ of a wave-speed interface are $0.948$, $0.441$, and
$0.049$, whereas the mean normalized residual magnitudes are $0.106$, $0.097$,
and $0.327$. Residual magnitude therefore does not preserve the gain ordering.
These results indicate that residual localization is a useful diagnostic for
closure benefit in this benchmark.

For the initial-condition-level correlation analysis, one statistical unit is
defined by an initial condition and a background sharpness value. The closure
error is first averaged over the three neural seeds within each unit, whereas
the structured error and $\Lambda_{0.1}$ are deterministic. The eight initial
conditions and three background sharpness values therefore give $n=24$
analysis units; repeated neural seeds are not treated as independent
observations. Across these units, $\Lambda_{0.1}$ and $\log G$ are strongly
associated: Pearson $r=0.847$ ($p=1.75\ee{-7}$, bootstrap 95\% CI
$[0.764,0.912]$) and Spearman $\rho_{\mathrm{S}}=0.887$
($p=7.80\ee{-9}$, bootstrap 95\% CI $[0.729,0.940]$). By contrast, normalized
residual magnitude and $\log G$ do not show a significant linear association
(Pearson $r=-0.293$, $p=0.165$).

\begin{table}[t]
\centering
\begin{tabular}{cccc}
\toprule
$s_{\mathrm{bg}}$ & Mean gain & Mean $\Lambda_{0.1}$ & Mean $\rho_i$ \\
\midrule
50 & $3.65$ & $0.948$ & $0.106$ \\
100 & $1.02$ & $0.441$ & $0.097$ \\
200 & $1.00$ & $0.049$ & $0.327$ \\
\bottomrule
\end{tabular}
\caption{Wave residual diagnostics across background sharpness levels. All
three reported quantities are averaged over the eight initial conditions.}
\label{tab:supp-wave-localization}
\end{table}

%% file: src/supp_reproducibility.tex
\section{Reproducibility Notes}

Reported error bars are standard deviations across the stated number of seeds.
The code release will include task configurations, execution scripts, reference-data
generation, metric extraction, and figure-generation scripts. Main neural
experiments use one NVIDIA A30 GPU with 24\,GB of device memory and two CPU
threads on a host with Intel Xeon Gold 6226R processors and 1\,TB of memory.
Deterministic CPU experiments and the standalone three-task cost benchmark use
two CPU threads on the same host. The software environment is Ubuntu 22.04.4
LTS with Linux 5.15.0-179, NVIDIA driver 580.159.03, Python 3.10.16,
PyTorch 2.5.1~\cite{paszke2019pytorch} built for CUDA 12.1, cuDNN 9.1.0,
and TorchDyn 1.0.6. No standalone CUDA compiler toolkit was used. The implementation uses
TorchDyn's NeuralODE interface~\cite{poli2021torchdyn} and explicit RK4
integration. The release will preserve the exact experiment directories used
to generate every reported table cell.

%% file: src/supp_visualizations.tex
\FloatBarrier
\begin{figure*}[t]
\centering
\includegraphics[width=\textwidth]{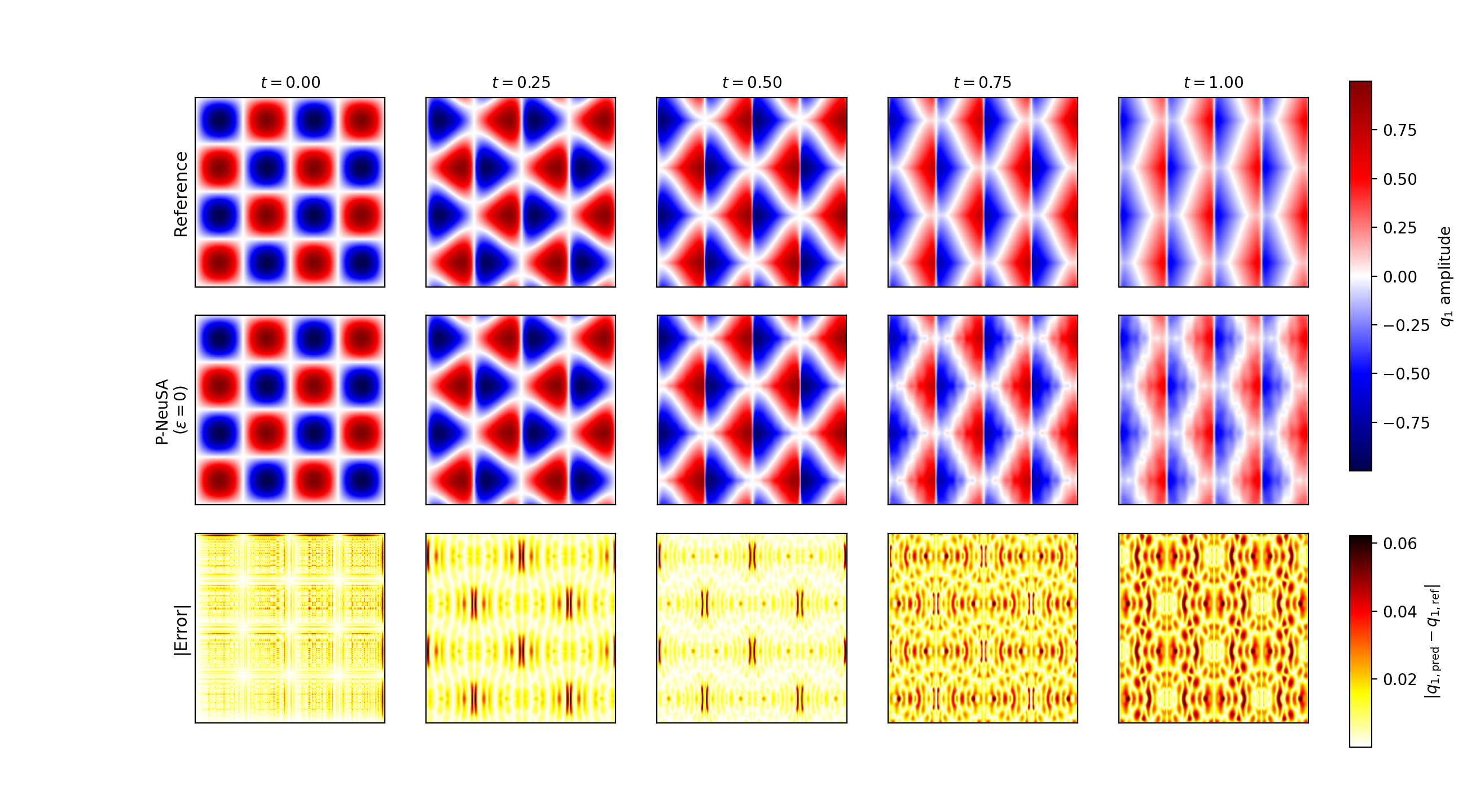}
\caption{2D Burgers $q_1$-component visualization for P-NeuSA
($\varepsilon=0$). The top row shows the reference solution, the middle row
shows the P-NeuSA prediction, and the bottom row shows
$|q_{1,\mathrm{pred}}-q_{1,\mathrm{ref}}|$ at each selected time.}
\label{fig:supp-burgers}
\end{figure*}

\begin{figure*}[t]
\centering
\includegraphics[width=\textwidth]{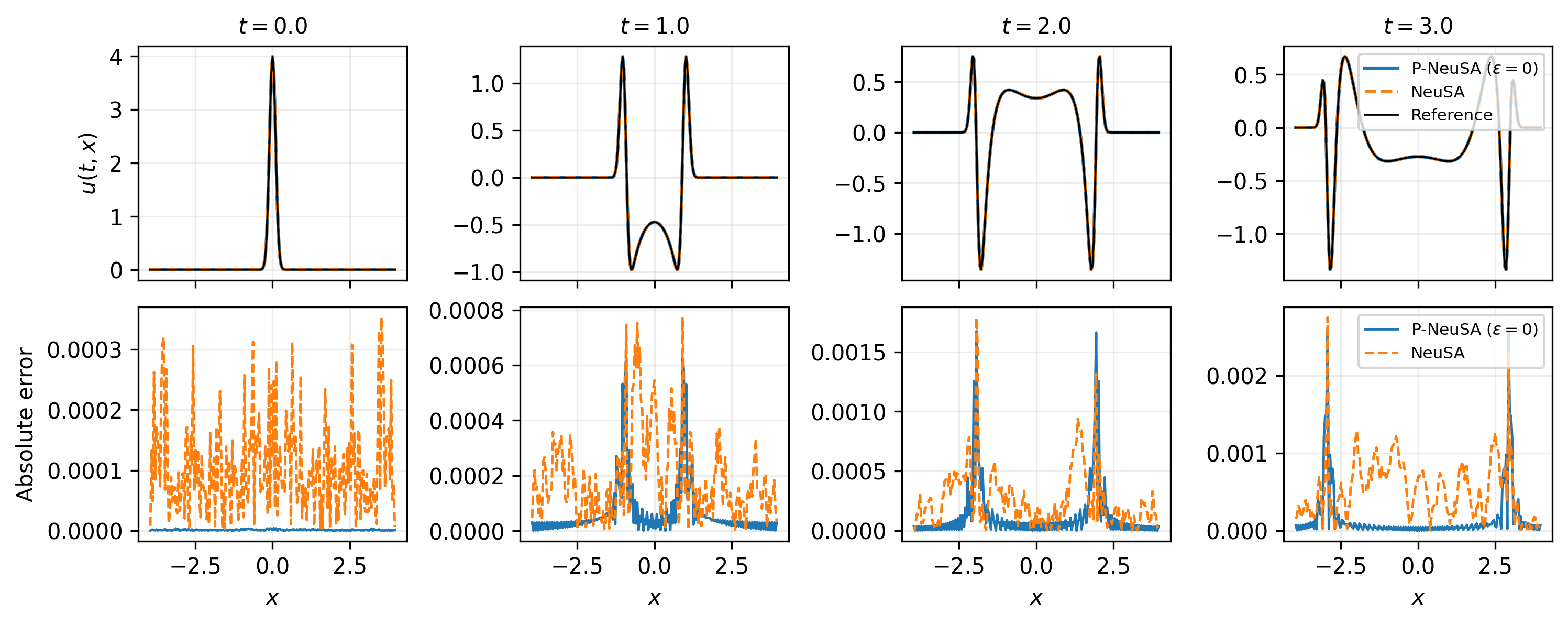}
\caption{Klein--Gordon comparison at selected time snapshots. The top row
compares P-NeuSA ($\varepsilon=0$), NeuSA, and the reference solution, while
the bottom row reports their pointwise absolute errors relative to the
reference.}
\label{fig:supp-kg}
\end{figure*}

\begin{figure*}[t]
\centering
\includegraphics[width=\textwidth]{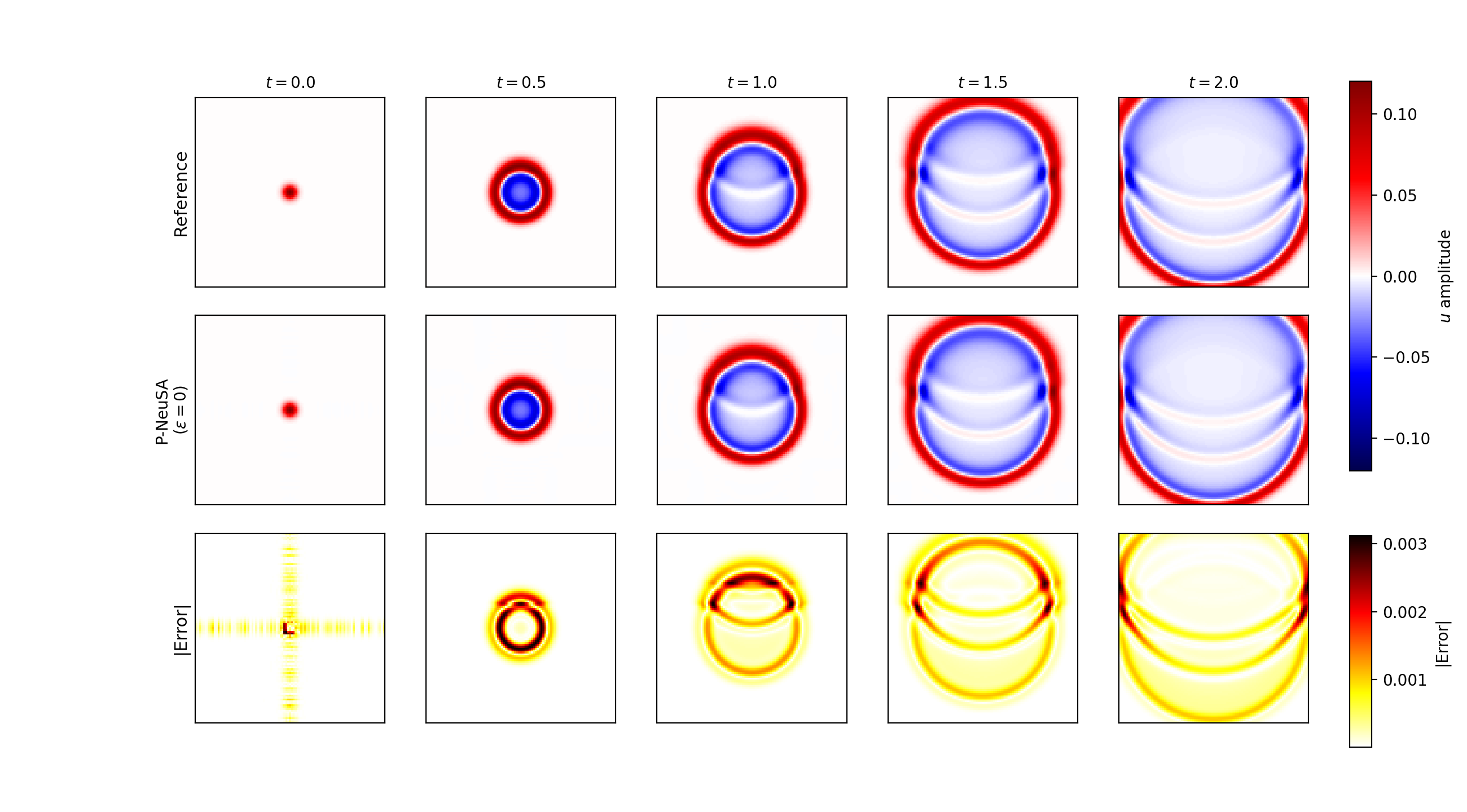}
\caption{Heterogeneous wave P-NeuSA ($\varepsilon=0$) visualization. The top
row shows the reference solution, the middle row shows the P-NeuSA prediction,
and the bottom row shows the pointwise absolute error at each selected time.}
\label{fig:supp-wave}
\end{figure*}